%% file: acl_latex.tex
\pdfoutput=1
\documentclass[11pt]{article}

\usepackage[final]{acl}

\usepackage{times}
\usepackage{latexsym}

\usepackage[T1]{fontenc}
\usepackage[utf8]{inputenc}

\usepackage{microtype}

\usepackage{inconsolata}

\usepackage{graphicx}

\usepackage{amsmath}
\usepackage{amssymb}
\usepackage{arydshln}
\usepackage{caption}
\usepackage{multirow}
\usepackage{subcaption}
\usepackage{comment}

\title{Improving Interpretability of Lexical Semantic Change with Neurobiological Features}

\author{Kohei Oda\textsuperscript{1}{\qquad}Hiroya Takamura\textsuperscript{2}{\qquad}Kiyoaki Shirai\textsuperscript{1}{\qquad}Natthawut Kertkeidkachorn\textsuperscript{1} \\
    \textsuperscript{1}Japan Advanced Institute of Science and Technology \\
    \textsuperscript{2}National Institute of Advanced Industrial Science and Technology \\
    \textsuperscript{1}\texttt{\{s2420017,kshirai,natt\}@jaist.ac.jp} \\
    \textsuperscript{2}\texttt{takamura.hiroya@aist.go.jp}}

\begin{document}

\maketitle

\begin{abstract}
Lexical Semantic Change (LSC) is the phenomenon in which the meaning of a word change over time. 
Most studies on LSC focus on improving the performance of estimating the degree of LSC, however, it is often difficult to interpret how the meaning of a word change. 
Enhancing the interpretability of LSC is a significant challenge as it could lead to novel insights in this field. 
To tackle this challenge, we propose a method to map the semantic space of contextualized embeddings of words obtained by a pre-trained language model to a neurobiological feature space. 
In the neurobiological feature space, each dimension corresponds to a primitive feature of words, and its value represents the intensity of that feature. 
This enables humans to interpret LSC systematically. 
When employed for the estimation of the degree of LSC, our method demonstrates superior performance in comparison to the majority of the previous methods. 
In addition, given the high interpretability of the proposed method, several analyses on LSC are carried out. 
The results demonstrate that our method not only discovers interesting types of LSC that have been overlooked in previous studies but also effectively searches for words with specific types of LSC.\footnote{Our code is available at: \url{https://github.com/iehok/LSC_with_Binder}.}
\end{abstract}

\begin{figure*}[t]
    \centering
    \includegraphics[width=1.0\linewidth]{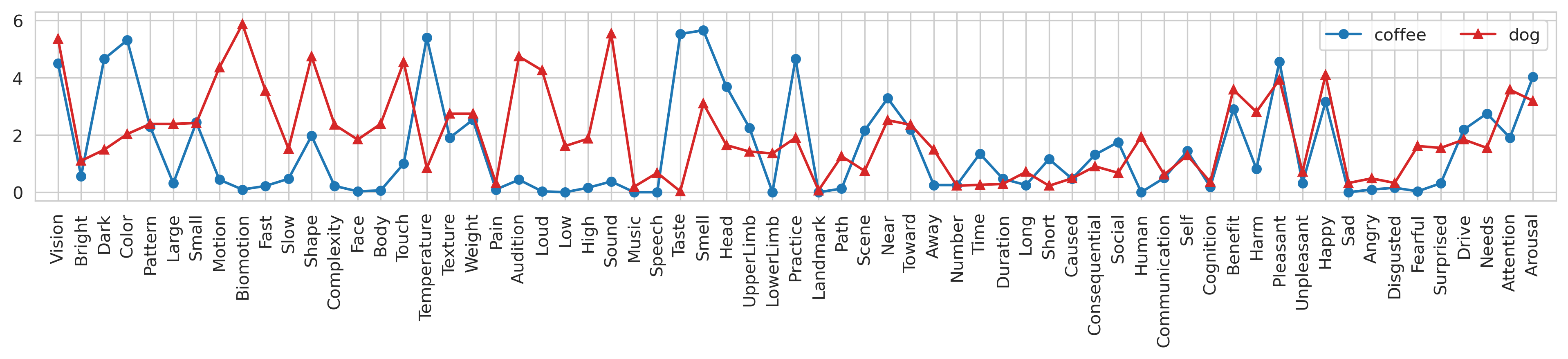}
    \caption{Binder feature values for ``coffee'' and ``dog''}
    \label{fig:binder_coffee_and_dog}
\end{figure*}

\section{Introduction}
The meanings of words change over time. 
For example, according to the Oxford English Dictionary (OED)\footnote{\url{https://www.oed.com/}}, the word \textit{gay} acquired the meaning of \textit{homosexual} around 1934, in addition to its earlier meaning of \textit{cheerful}. 
This phenomenon is called Lexical Semantic Change (LSC) and actively studied in recent years \citep{tahmasebi2019surveycomputationalapproacheslexical, Tahmasebi2021, 10.1145/3672393}. 
Many studies in this field represent the meanings of words as vectors using embedding models, such as static word embeddings \citep{mikolov2013efficientestimationwordrepresentations} and BERT \citep{devlin-etal-2019-bert}, and learn separate spaces for different time periods \citep{kim-etal-2014-temporal, hamilton-etal-2016-diachronic, 10.5555/3305381.3305421} or handle multiple time periods within the same space \citep{hu-etal-2019-diachronic, giulianelli-etal-2020-analysing, 10.1145/3366424.3382186}. 
While these techniques are useful for estimating the degree of LSC, they are inappropriate for humans to interpret LSC. 

Several methods have been proposed to improve the interpretability of LSC, including a method presenting neighboring words in a vector space \citep{gonen-etal-2020-simple}, obtaining representative co-occurrence words \citep{montariol-etal-2021-scalable}, predicting substitutions \citep{card-2023-substitution}, assigning predefined word senses \citep{tang-etal-2023-word}, and generating definition sentences of word meanings \citep{giulianelli-etal-2023-interpretable, fedorova-etal-2024-definition}. 
These methods help humans interpret LSC through natural language, e.g., by showing indicative words or definition sentences. 
However, explanations of LSC based on words and sentences are ambiguous and lack a systematic explanatory framework. 

Motivated by the above, we propose a method to improve the interpretability of LSC by using neurobiological features proposed by \citet{Binder18052016}, which we call \textit{Binder features} in this paper.
There are 65 Binder features such as \textit{Vision}, \textit{Audition}, and \textit{Happy}.
The values of these 65 Binder features have been estimated for 535 English words and are open to the public.\footnote{\url{https://www.neuro.mcw.edu/index.php/resources/brain-based-semantic-representations/}}
Based on previous studies \citep{utsumi2018neurobiologicallymotivatedanalysisdistributional, https://doi.org/10.1111/cogs.12844, turton-etal-2021-deriving}, we use the public dataset above to train a regression model that maps the BERT semantic space to the Binder space for the quantitative and multi-perspective interpretation of LSC. 

First, the potential of the Binder features in analyzing LSC is evaluated by applying our method to a task aimed at estimating the degree of LSC. 
Second, utilizing the high interpretability of our method, we analyze types of LSC. 
The integration of our method with Sparse PCA (Principal Component Analysis) enables us to identify interesting types of LSC that have not been found in previous studies. 
Finally, we apply our method to detect amelioration and pejoration \citep{1360013171975918848}, and successfully identify ameliorative and pejorative words in a real corpus. 

The contributions of our paper are summarized as follows: 
\begin{itemize}
    \item We introduce neurobiological features into the field of lexical semantic change, thereby improving the interpretability. 
    \item We discover several interesting types of LSC that have not been noted in previous studies by combining our method with Sparse PCA. 
    \item We propose a method that can easily detect specific types of LSC, amelioration and pejoration, using our approach. 
\end{itemize}

\section{Related Work}

\subsection{Lexical Semantic Change}

LSC is mainly studied in the fields of linguistics and natural language processing (NLP). 
Even when being constrained to NLP, numerous methods are proposed such as a method that utilizes mutual information \citep{gulordava-baroni-2011-distributional, hamilton-etal-2016-diachronic, schlechtweg-etal-2019-wind}, Bayesian models \citep{emms-jayapal-2016-dynamic, frermann-lapata-2016-bayesian, inoue-etal-2022-infinite}, and static word embeddings \citep{10.1145/2736277.2741627, takamura-etal-2017-analyzing, del-tredici-etal-2019-short}. 

Recently, with the emergence of pre-trained language models such as BERT \citep{devlin-etal-2019-bert} and RoBERTa \citep{liu2019robertarobustlyoptimizedbert} that can generate representations of the meanings of words in a context, methods using these models have been actively studied \citep{kutuzov-giulianelli-2020-uio, martinc-etal-2020-leveraging, liu-etal-2021-statistically}. 
\citet{hu-etal-2019-diachronic} propose a method to identify how the meaning of a word changes by calculating the distribution of word senses over time, where example sentences in the OED are used to assign senses to words in the corpus. 
\citet{giulianelli-etal-2020-analysing} propose a method to calculate the distribution of usage types (pseudo senses) without using a dictionary. 
This method uses $k$-means clustering on a set of contextualized embeddings from all time periods to assign usage types to words in the corpus. 
Additionally, the degree of LSC between two different time periods is estimated using either the Jensen-Shannon divergence (JSD) between usage type distributions or the average pairwise distance (APD) between sets of contextualized embeddings from these time periods. 
In this study, we model LSC based on \citet{giulianelli-etal-2020-analysing} coupled with the Binder features to improve the interpretability. 

\subsection{Interpretable Word Embeddings}

Interpretable representations, i.e., methods of assigning roles (interpretations) to each dimension of an embedding, have been extensively studied \citep{panigrahi-etal-2019-word2sense, _enel_2020, aloui-etal-2020-slice}. 
However, these methods often suffer from a lack of clarity of a role for each dimension or coarse granularity of roles. 

\citet{Binder18052016} propose interpretable word vectors by defining 65 features based on neurobiological perspectives and manually assign these strengths (0 to 6) to 535 words, including 434 nouns, 62 verbs, and 39 adjectives. 
Figure \ref{fig:binder_coffee_and_dog} shows the Binder features and their corresponding values for ``coffee'' and ``dog.'' 
The values of \textit{Taste} and \textit{Smell} are relatively high for ``coffee,'' while the values of \textit{Biomotion} and \textit{Sound} are high for ``dog.'' 
Additionally, the \textit{Vision} feature has high values for both words. 

Binder features are actively studied in the fields of cognitive linguistics and NLP. 
\citet{https://doi.org/10.1111/cogs.12844, chersoni-etal-2021-decoding, flechas-manrique-etal-2023-enhancing} investigate what kind of information is encoded in static word embeddings, such as SGNS \citep{mikolov2013efficientestimationwordrepresentations} and GloVe \citep{pennington-etal-2014-glove}, by mapping these word embedding spaces to the Binder feature space. 
\citet{turton-etal-2020-extrapolating} assign the Binder values to words other than the original 535 words by the aforementioned mapping from word embeddings. 
\citet{turton-etal-2021-deriving} demonstrate that contextualized word embeddings generated from Transformer \citep{NIPS2017_3f5ee243} based models such as BERT \citep{devlin-etal-2019-bert} and RoBERTa \citep{liu2019robertarobustlyoptimizedbert} can derive the Binder values in the same way as the mapping of static word embeddings. 

\section{Mapping BERT Space to Binder Space}
\label{sec:mapping_bert_space_to_binder_space}

To enhance the interpretability of LSC, we first convert the semantic space of contextualized word embeddings derived from BERT to that of the Binder features. 
Specifically, a regression model is trained, which maps the BERT space (768 dimensions) to the Binder space (65 dimensions). 
The regression model, designated as $\psi$, is formalized as follows, 
\begin{equation}
    \mathbf{b}_w = \psi(\mathbf{r}_w), 
\end{equation}
where $\mathbf{r}_w$ and $\mathbf{b}_w$ are BERT and Binder vectors, respectively. 

\subsection{Word Embeddings on BERT Space}
Let $\mathcal{C}$ be the corpus used for training, and let $\mathcal{C}_w$ be the set of $(s, i)$, a pair of a sentence $s$ in $\mathcal{C}$ that contains the target word $w$ and its position $i$ in $s$. 
The representation of $w$ in the entire $\mathcal{C}$ is defined as follows. 
\begin{equation}
    \mathbf{r}_w = \frac{1}{|\mathcal{C}_w|}\sum_{(s, i) \in \mathcal{C}_w}{\phi}(s, i). 
\end{equation}
The function ${\phi}(s, i)$ denotes the hidden state of the final layer for the $i$-th token of the BERT model when $s$ is given as an input. 
In this study, \texttt{bert-base-uncased}\footnote{\url{https://huggingface.co/google-bert/bert-base-uncased}} is used as the BERT model. 

The Clean Corpus of Historical American English (CCOHA) \citep{alatrash-etal-2020-ccoha} is used as $\mathcal{C}$. 
CCOHA is an English corpus covering the period from 1820 to 2020, divided into ten-year segments. 
It consists of five genres: TV/Movies, Fiction, Magazine, Newspaper, and Non-fiction. 
% Following \citet{giulianelli-etal-2020-analysing}, the text from 1910 to 2010 is utilized to obtain $\mathbf{r}_w$. 

\subsection{Training of Regression Model}
Two architectures of the regression model are applied: a simple linear transformation (LT) and a multilayer perceptron (MLP). 
The MLP consists of four hidden layers (300, 200, 100, 50 dimensions), following \citet{turton-etal-2021-deriving}. 
The output of each layer is activated by ReLU. 
To match the scale of the Binder value, in both the LT and the MLP, the final output is activated by Sigmoid and subsequently multiplied by 6 to convert values within the range of 0 to 6. 
The regression models are trained using 535 words associated with the Binder feature vectors \citep{Binder18052016}. 
The loss function is set to the mean squared error (MSE) between predicted and ground-truth values of all Binder features of the target words. 

\subsection{Settings}
% Given that the Binder feature vectors were created in the 2010s, it is hypothesized that a better regression model is obtained when a corpus from the 2010s is used. 
We conduct experiments using two different periods of CCOHA: 1910-2010 and 1960-2010. 
The period 1910-2010 follows the setting in \citet{giulianelli-etal-2020-analysing}, while the period 1960-2010 is set to the most recent half of it, as the Binder dataset \citep{Binder18052016} was created in 2016. 
The performance of the trained regression model is evaluated using $k$-fold cross-validation, where $k$ is set to 10. 
The batch size, the learning rate, and the number of epochs are set to 16, 1e-3 and 100, respectively. 
The quality of the regression model is evaluated by the MSE on the test set. 
The MSE is measured at each epoch, and the minimum MSE is recorded. 

\begin{table}[t]
    \centering
    \begin{tabular}{c|cc}
        \hline
                    & LT        & MLP   \\
        \hline
        1910-2010   & .571          & .645  \\
        1960-2010   & \textbf{.569} & .689  \\
        \hline
    \end{tabular}
    \caption{Average MSE for 10 trials}
    \label{tab:result_k_fold_cross_validation}
\end{table}

\subsection{Results}
Table \ref{tab:result_k_fold_cross_validation} shows the average MSE for 10 trials of the cross-validation. 
LT significantly outperforms MLP, while the time period of the training corpus has a minimal influence on the results. 
This may be because the words in the Binder dataset are well-known and common, which leads to a relatively stable representation over time. 
This finding partially agrees with the results obtained by \citet{hamilton-etal-2016-diachronic}. 

\section{Lexical Semantic Change Detection}
\label{sec:evaluation_with_semeval_2020_task_1}

This section proposes and evaluates a method to detect LSC using the Binder feature vectors. 

\subsection{Task Definition}
SemEval-2020 Task 1 (Subtask 2) \citep{schlechtweg-etal-2020-semeval} is a task that aims to predict the degree of LSC of a word $w$. 
Specifically, the goal is to predict an LSC score representing how drastically the meaning of $w$ changes between $\mathcal{C}^{t_1}$ and $\mathcal{C}^{t_2}$, which are corpora of two different periods $t_1$ and $t_2$. 
The dataset consists of 37 English target words with manually assigned LSC scores. 
$\mathcal{C}^{t_1}$ and $\mathcal{C}^{t_2}$ are parts of CCOHA from 1810 to 1860 and 1960 to 2010, respectively. 
Evaluation is performed by measuring Spearman's rank correlation coefficient between the predicted and ground-truth LSC scores. 

\subsection{Predicting Degree of LSC}
To predict the degree of LSC of a word $w$ from $t_1$ to $t_2$, the set of contextualized embeddings of $w$ in the corpus $\mathcal{C}^t_w$ is calculated for each time period: 
\begin{equation}
    \mathcal{U}^t_w = \bigcup_{(s, i) \in \mathcal{C}^t_w}\{\,\psi(\phi(s, i))\,\}, 
\end{equation}
where $\psi$ is the regression model, either LT or MLP explained in Section \ref{sec:mapping_bert_space_to_binder_space}. 
Then, following \citet{giulianelli-etal-2020-analysing}, the degree of LSC between $\mathcal{U}^{t_1}_w$ and $\mathcal{U}^{t_2}_w$ is measured by the average pairwise distance (APD): 
\begin{equation}
\begin{aligned}
    \mathrm{APD}&(\mathcal{U}^{t_1}_w, \mathcal{U}^{t_2}_w) = \\
    &\frac{1}{|\mathcal{U}^{t_1}_w| \cdot |\mathcal{U}^{t_2}_w|}\sum_{\mathbf{u}_i \in \mathcal{U}^{t_1}_w}\sum_{\mathbf{u}_j \in \mathcal{U}^{t_2}_w}d(\mathbf{u}_i, \mathbf{u}_j), 
\end{aligned}
\end{equation}
where $d$ is a distance function. 
We compared the following three distance functions: Euclidean distance, cosine distance, and Spearman distance. 
Spearman distance is defined as $(1 - \mathrm{sc}(\mathbf{u}_i,\mathbf{u}_j))$, where $\mathrm{sc}(\mathbf{u}_i,\mathbf{u}_j)$ is Spearman's rank correlation coefficient between sets of values of dimensions in two vectors. 

\begin{table}[t]
    \centering
    \setlength{\tabcolsep}{5pt}
    \begin{tabular}{l|ccc}
        \hline
        Model           & Euclid        & Cosine        & Spearman      \\
        \hline
        BERT space      & .616          & .645          & .618          \\
        LT-1910-2010    & .633          & .644          & \textbf{.647} \\
        LT-1960-2010    & \textbf{.635} & \textbf{.667} & .634          \\
        MLP-1910-2010   & .499          & .587          & .562          \\
        MLP-1960-2010   & .483          & .442          & .540          \\
        \hline
    \end{tabular}
    \caption{Spearman's rank correlation coefficient for SemEval-2020 Task 1. ``BERT space'' means a method that calculates APD in the BERT space without mapping it to the Binder space.}
    \label{tab:result_semeval_2020_1}
\end{table}

\begin{table}[t]
    \centering
    \setlength{\tabcolsep}{3pt}
    \begin{tabular}{lcc}
        \hline
        Model                                       & EK            & Score         \\
        \hline
        SSCS \citep{tang-etal-2023-word}            & \checkmark    & .589          \\
        XL-LEXEME \citep{cassotti-etal-2023-xl}     & \checkmark    & .757          \\
        SDML \citep{aida-bollegala-2024-semantic}   & \checkmark    & .774          \\
        \hline
        NLPCR \citep{rother-etal-2020-cmce}         &               & .436          \\
        APD \citep{laicher-etal-2021-explaining}    &               & .571          \\
        ScaledJSD \citep{card-2023-substitution}    &               & .547          \\
        SSCD \citep{aida-bollegala-2023-swap}       &               & .383          \\
        LT-1960-2010 (ours)                     &               & \textbf{.667} \\
        \hline
    \end{tabular}
    \caption{Spearman's rank correlation coefficient for previous methods on SemEval-2020 Task 1. ``EK'' (external knowledge) means methods that are fine-tuned with WiC corpora \citep{raganato-etal-2020-xl, martelli-etal-2021-semeval, liu-etal-2021-am2ico} or methods using the information of dictionaries such as WordNet \citep{miller-1994-wordnet} and BabelNet \citep{navigli-ponzetto-2010-babelnet}.}
    \label{tab:result_semeval_2020_2}
\end{table}

\subsection{Results}
Table \ref{tab:result_semeval_2020_1} shows a comparison between the baseline BERT space and our methods based on different regression models. 
The performance of LSC detection is slightly improved by mapping to the Binder space using the linear regression model. 
The architecture of the regression model significantly impacts the performance of LSC detection, while the period of the corpus used for training the regression model has a relatively small impact; this tendency is similar to that in Table \ref{tab:result_k_fold_cross_validation}. 
Among the three distance functions, the cosine distance is relatively stable and performs well. 

Table \ref{tab:result_semeval_2020_2} shows a comparison of our method with other existing methods. 
Although our method is simple, it achieves the best performance compared to other methods that do not use any external knowledge. 

\begin{figure*}[t]
    \centering
    \includegraphics[width=1.0\linewidth]{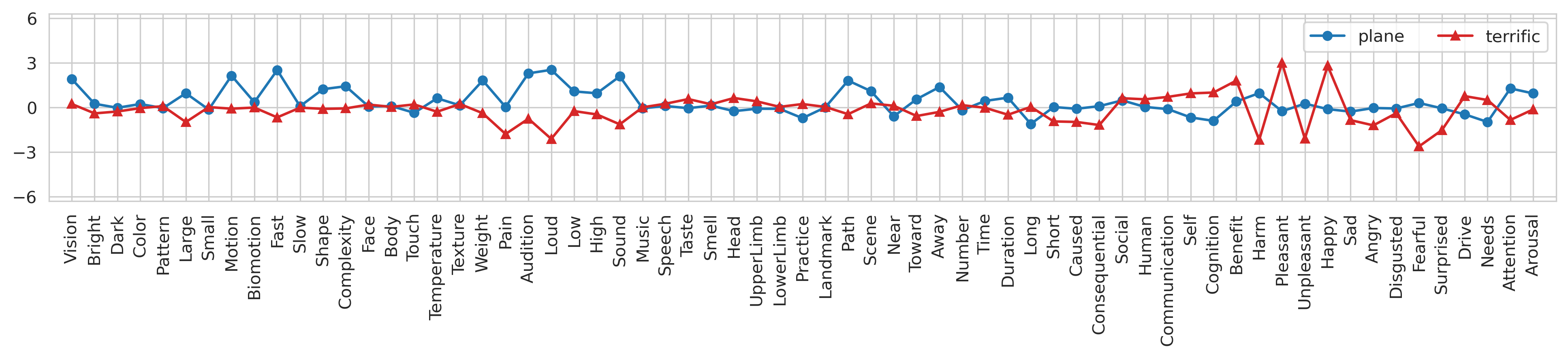}
    \caption{LSC vectors for \textit{plane} and \textit{terrific}}
    \label{fig:lsc_plane_and_terrific}
\end{figure*}

\section{Analysis of LSC Types}
\label{sec:analyzing_tyes_of_lsc}

This section describes an analysis of LSC types using our method. 
Given the high interpretability of neurobiological features, our goal is to identify the types of semantic changes of words between two different periods $t_1$ and $t_2$. 

\subsection{Target Words}
\label{subsec:analyzing_tyes_of_lsc-target_words}
The target words used for this analysis are collected from WordNet \citep{miller-1994-wordnet} lemmas meeting the following three conditions: (i) included in the vocabulary of the BERT tokenizer, (ii) having two or more senses in WordNet, and (iii) four or more characters long. 
Condition (i) is set because our method is incapable of handling words that are divided into subwords. 
Condition (ii) is set because words whose meanings have changed are likely to have newly added senses, thereby resulting in polysemy. 
Condition (iii) is set because short words are more likely to become subwords within other words. 
Based on these three conditions, a total of 8,570 target words are chosen. 

\subsection{Corpora}
The CCOHA 1910s (from 1910 to 1920) and 2000s (from 2000 to 2010) are used as the corpora $\mathcal{C}^{t_1}$ and $\mathcal{C}^{t_2}$. 
This corresponds to the first and last decades of the period from 1910 to 2010 used for training the regression model (Section \ref{sec:mapping_bert_space_to_binder_space}). 

\subsection{Methods}
To analyze how the meaning of words changed between two periods $t_1$ and $t_2$, the LSC vector of the word $w$, denoted as $\mathbf{v}_\mathrm{lsc}(w)$, is computed as follows: 
\begin{equation}
\begin{aligned}
    \mathbf{v}_\mathrm{lsc}(w)\!=\!\frac{1}{|\mathcal{U}^{t_2}_w|}\sum_{\mathbf{u}_i \in \mathcal{U}^{t_2}_w}{\mathbf{u}_i}\!-\!\frac{1}{|\mathcal{U}^{t_1}_w|}\sum_{\mathbf{u}_i \in \mathcal{U}^{t_1}_w}{\mathbf{u}_i}. 
\end{aligned}
\end{equation}
This vector represents the semantic changes of all Binder features. 
A positive value in a dimension of the LSC vector means that the meaning of the corresponding Binder feature is newly acquired from $t_1$ to $t_2$, while a negative value implies a loss of the meaning. 

After calculating the LSC vectors for all target words, Sparse PCA is applied to the LSC vectors of the 500 target words with the largest norms, supposing that the meanings of words with small norms are not significantly changed. 
Unlike conventional PCA, Sparse PCA enhances interpretability by setting many elements in the eigenvectors to zero, and the eigenvectors do not need to be orthogonal to each other. 
Since the number of principal components should be predetermined, it is set to 10 in this experiment. 
The analysis of different numbers of principal components remains a subject for future work. 

It is hypothesized that each principal component (PC) of Sparse PCA represents a type of LSC. 
For each PC, the top three Binder features with the highest values in the eigenvector are extracted to provide a clear interpretation of the LSC type. 
Subsequently, we check the words in descending or ascending order of their values of the principal component and verify whether they are representative words. 
The validity of the chosen representative words is evaluated by the following procedures. 
First, following \citet{giulianelli-etal-2020-analysing}, usage types (pseudo senses) are assigned to the target words in example sentences by conducting $k$-means clustering on a set of contextualized embeddings. 
Second, the five examples closest to the center of each cluster are examined to confirm whether they are correctly divided according to their meanings. 
Finally, the change in the distribution of usage types from $t_1$ to $t_2$ is checked to investigate whether it supports the LSC type augmented by the related Binder features. 

This method is similar to the analysis by applying PCA in BERT space \citep{aida-bollegala-2025-investigating}, but enhances the interpretability of LSC types. 
First, not only words with large or small principal components but also the values of eigenvectors can be used for analyzing LSC types. 
Second, since Sparse PCA assigns a zero to many elements, it is easier to find relevant (non-zero) Binder features for each LSC type. 

\begin{figure*}[t]
    \centering
    \includegraphics[width=1.0\linewidth]{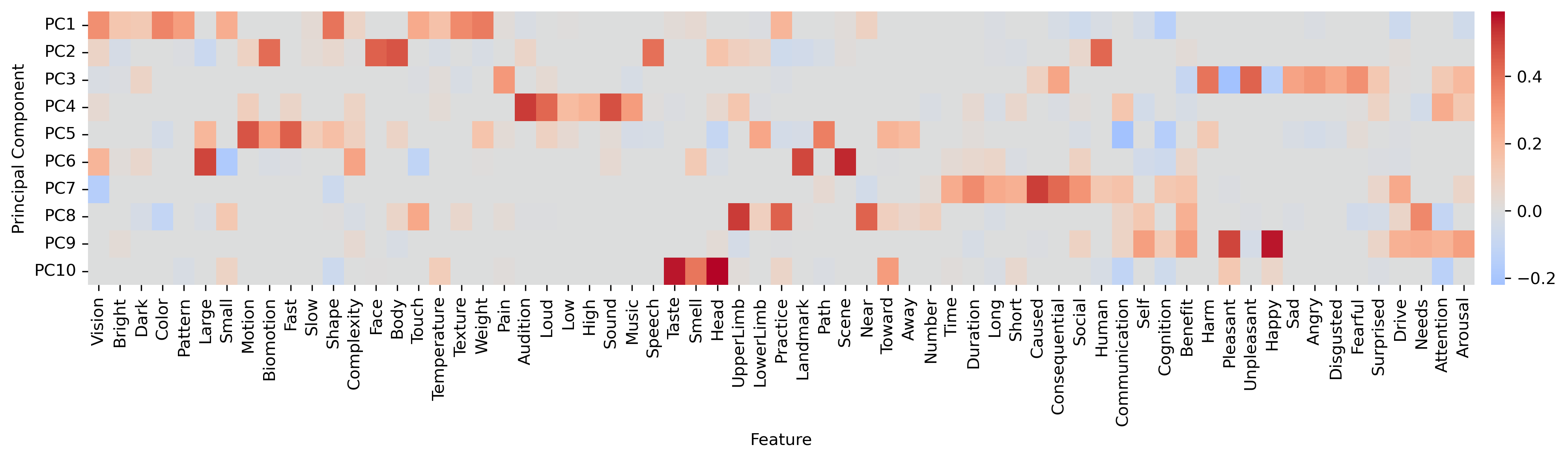}
    \caption{Eigenvectors obtained by Sparse PCA. The number of principal components is set to 10. The horizontal and vertical axes represent the 65 Binder features and the 10 principal components, respectively.}
    \label{fig:pca}
\end{figure*}

% \begin{figure}[t]
%     \centering
%     \includegraphics[width=0.8\linewidth]{figures/cevr.png}
%     \caption{Cumulative explained variance ratio}
%     \label{fig:cevr}
% \end{figure}

\begin{table*}[!t]
    \centering
    \begin{tabular}{r|l|l|l}
        \hline
        PC                  & LSC Type Label                    & Top 3 Binder Features                             & Representative Words \\
        \hline
        \hline
        \multirow{2}{*}{1}  & \multirow{2}{*}{Artifact}         & \multirow{2}{*}{Shape, Weight, Color}             & $\uparrow$ console, plastic, vogue \\
                            &                                   &                                                   & $\downarrow$ overall, album, bluegrass \\
        \hline
        \multirow{2}{*}{2}  & \multirow{2}{*}{Human}            & \multirow{2}{*}{Body, Face, Human}                & $\uparrow$ coach, shooter, racer \\
                            &                                   &                                                   & $\downarrow$ yahoo, explorer, major \\
        \hline
        \multirow{2}{*}{3}  & \multirow{2}{*}{Negative Meaning} & \multirow{2}{*}{Unpleasant, Harm, Fearful}        & $\uparrow$ serial, aids, parkinson \\
                            &                                   &                                                   & $\downarrow$ offence, terrific, crook \\
        \hline
        \multirow{2}{*}{4}  & \multirow{2}{*}{Sound}            & \multirow{2}{*}{Audition, Sound, Loud}            & $\uparrow$ bluegrass, plane, blues \\
                            &                                   &                                                   & $\downarrow$ instrumentation, click, booming \\
        \hline
        \multirow{2}{*}{5}  & \multirow{2}{*}{Transportation}   & \multirow{2}{*}{Motion, Fast, Path}               & $\uparrow$ pickup, sedan, plane \\
                            &                                   &                                                   & $\downarrow$ steamed, omnibus, coach \\
        \hline
        \multirow{2}{*}{6}  & \multirow{2}{*}{Place}            & \multirow{2}{*}{Scene, Large, Landmark}           & $\uparrow$ facility, resort, berkeley \\
                            &                                   &                                                   & $\downarrow$ manila, chihuahua, bologna \\
        \hline
        \multirow{2}{*}{7}  & \multirow{2}{*}{Social}           & \multirow{2}{*}{Caused, Consequential, Duration}  & $\uparrow$ warming, launch, summit \\
                            &                                   &                                                   & $\downarrow$ briefs, console, offensive \\
        \hline
        \multirow{2}{*}{8}  & \multirow{2}{*}{Familiar Thing}   & \multirow{2}{*}{UpperLimb, Practice, Near}        & $\uparrow$ topical, sink, blackberry \\
                            &                                   &                                                   & $\downarrow$ album, shooter, warming \\
        \hline
        \multirow{2}{*}{9}  & \multirow{2}{*}{Positive Meaning} & \multirow{2}{*}{Happy, Pleasant, Benefit}         & $\uparrow$ bonding, outgoing, terrific \\
                            &                                   &                                                   & $\downarrow$ intelligence, utility, console \\
        \hline
        \multirow{2}{*}{10} & \multirow{2}{*}{Food}             & \multirow{2}{*}{Head, Taste, Smell}               & $\uparrow$ bologna, bourbon, steamed \\
                            &                                   &                                                   & $\downarrow$ alcoholic, blackberry, bluegrass \\
        \hline
    \end{tabular}
    \caption{The result of analysis of Sparse PCA. ``LSC Type Label'' is a manually assigned label for the LSC type. The symbols $\uparrow$ and $\downarrow$ indicate words with relatively large and small principal component values, respectively, suggesting the words have acquired or lost their meanings of the features.}
    \label{tab:lsc_types}
\end{table*}

\begin{figure*}[t]
    \centering
    \begin{subfigure}[t]{0.329\linewidth}
        \includegraphics[width=\linewidth]{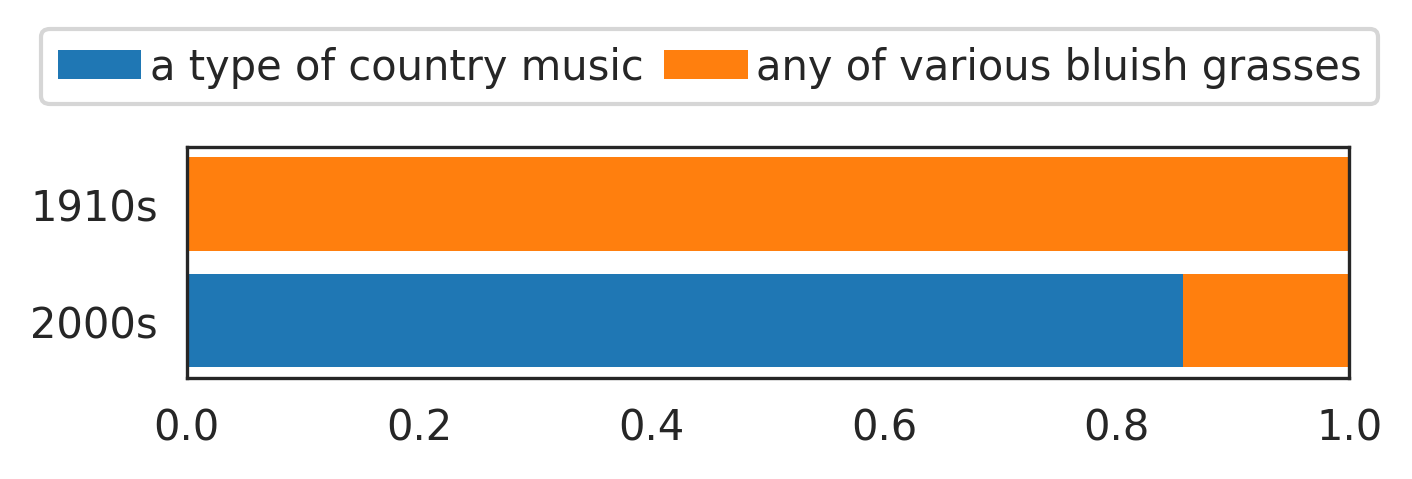}
        \caption{bluegrass}
        \label{fig:sense_dist_bluegrass}
    \end{subfigure}
    \begin{subfigure}[t]{0.329\linewidth}
        \includegraphics[width=\linewidth]{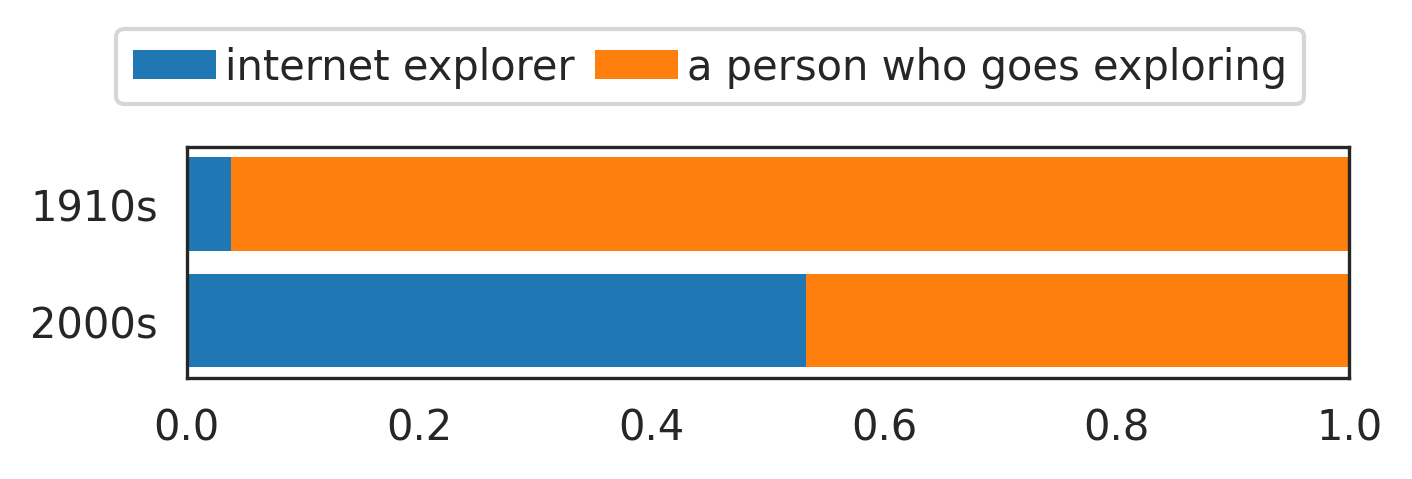}
        \caption{explorer}
        \label{fig:sense_dist_explorer}
    \end{subfigure}
        \begin{subfigure}[t]{0.329\linewidth}
        \includegraphics[width=\linewidth]{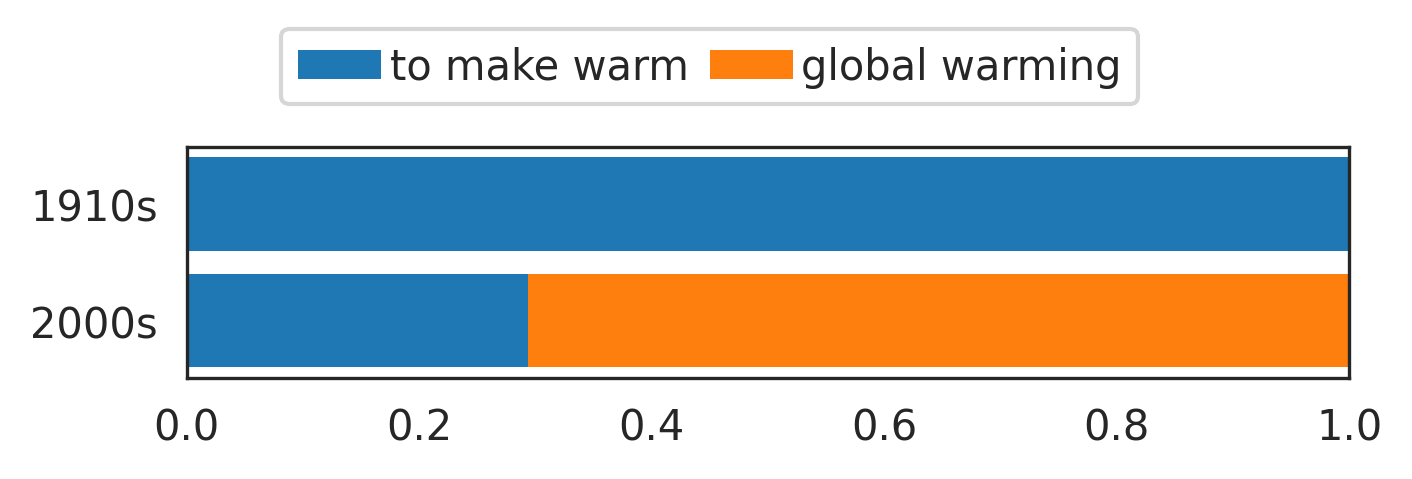}
        \caption{warming}
        \label{fig:sense_dist_warming}
    \end{subfigure}
    \caption{Distributions of the usage types for \textit{bluegrass}, \textit{explorer}, and \textit{warming}}
    \label{fig:sense_dist}
\end{figure*}

\subsection{Results}
\label{subsec:analyzing_tyes_of_lsc-results}

Figure \ref{fig:lsc_plane_and_terrific} shows the LSC vectors for \textit{plane} and \textit{terrific}. 
According to the OED, the word \textit{plane} acquired the meaning of \textit{airplane} around 1908, in addition to its existing meaning of \textit{a flat geometrical surface}. 
As illustrated in Figure \ref{fig:lsc_plane_and_terrific}, the values of the Binder features such as \textit{Motion}, \textit{Audition}, and \textit{Path} exhibit a substantial increase. 
Additionally, according to the OED, \textit{terrific} acquired the meaning of \textit{amazing} around 1871, in addition to its existing meaning of \textit{causing terror}. 
The values of the Binder features \textit{Pleasant} and \textit{Happy} have increased significantly, while the values of the Binder features \textit{Harm}, \textit{Unpleasant}, and \textit{Fearful} have decreased significantly. 
This indicates that the major meaning of \textit{terrific} has shifted from a negative to a positive meaning. 

Figure \ref{fig:pca} shows the eigenvectors obtained by Sparse PCA. 
Many elements in the eigenvectors are zero, making them relatively easy to interpret. 
In addition, by examining the absolute values in the eigenvectors, it is possible to identify Binder features that are deeply related to or not related to LSC. 
For example, the absolute values of \textit{Vision} at the first, sixth, and seventh PCs are relatively high, indicating \textit{Vision} is likely to be deeply related to LSC. 
On the other hand, the absolute values of \textit{Number} are nearly zero in all PCs, suggesting that \textit{Number} does not contribute to LSC. 

% Figure \ref{fig:cevr} shows the cumulative explained variance ratio (CEVR) for the set of vectors applied to Sparse PCA. 
% The subtle inclination in the curvature of CEVR shows that all PCs, not a limited number of them, are endowed with information concerning LSC. 

Table \ref{tab:lsc_types} shows the types of LSC corresponding to PCs. 
Many interesting types of LSC, which have not been noted in previous studies \citep{1360013171975918848, Campbell+2020}, are discovered by our method. 
Figure \ref{fig:sense_dist} shows the distributions of the usage types for some illustrative examples of words: \textit{bluegrass}, \textit{explorer}, and \textit{warming}. 
For \textit{bluegrass} in PC4, the meaning has changed from \textit{any of various bluish grasses} to \textit{a type of country music}, shifting to a meaning related to sounds. 
For \textit{explorer} in PC2, the meaning related to humans has declined due to its increased use in the collocation \textit{internet explorer}. 
For \textit{warming} in PC7, the meaning has changed to a meaning related to social due to its increased use in the collocation \textit{global warming}. 
The distributions for other words are shown in the Appendix \ref{sec:distributions_usage_types}. 

\section{Analysis of Amelioration and Pejoration}
\label{sec:analyzing_amelioration_and_pejoration}

The process of mapping the BERT space to the Binder space not only improves the interpretability of LSC, as described in Section \ref{sec:analyzing_tyes_of_lsc}, but also facilitates the search for words corresponding to specific types of LSC. 
This section presents a case study to search for words that went through amelioration or pejoration, where amelioration means acquiring positive sentiment and pejoration means acquiring negative sentiment \citep{1360013171975918848}. 
% In fact, amelioration and pejoration are identified as PC9 and PC3 in Table \ref{tab:lsc_types}. 
In addition, we evaluate the ability of our method to identify specific words that acquire a positive or negative meaning over time. 
 
\subsection{Known Words of Amelioration and Pejoration}

Several pieces of literature have already reported examples of amelioration and pejoration. 
From these references, the sets of known words of amelioration and pejoration, $\mathcal{W}_\mathrm{ame}$ and $\mathcal{W}_\mathrm{pej}$ respectively, are extracted. 
Table \ref{tab:known_words_amelioration_and_pejoration} shows $\mathcal{W}_\mathrm{ame}$ and $\mathcal{W}_\mathrm{pej}$ with their references. 
Although both sets are small, they are used as ground truth to examine whether our method successfully identifies these words as amelioration or pejoration.

\begin{table}[t]
    \centering
    \begin{tabular}{|c|l|}
        \hline
        \multirow{4}{*}{$\mathcal{W}_\mathrm{ame}$} & \textbf{hysteria} \citep{cook-stevenson-2010-automatically} \\
        \cdashline{2-2}
        & \textbf{brilliant}, \textbf{fabulous}, \textbf{fantastic}, \\
        & \textbf{spectacular} \citep{Altakhaineh_2018} \\
        \cdashline{2-2}
        & \textbf{terrific} \citep{deWit_2021} \\
        \hline
        \multirow{6}{*}{$\mathcal{W}_\mathrm{pej}$} & \textbf{dynamic}, \textbf{synthesis} \\
        & \citep{cook-stevenson-2010-automatically} \\
        \cdashline{2-2}
        & \textbf{abuse}, \textbf{addiction}, \textbf{harassment}, \\
        & \textbf{prejudice}, \textbf{trauma} \citep{Haslam02012016} \\
        \cdashline{2-2}
        & \textbf{terrible} \citep{Altakhaineh_2018} \\
        \cdashline{2-2}
        & \textbf{awful} \citep{deWit_2021} \\
        \hline
    \end{tabular}
    \caption{Sets of known words of amelioration $\mathcal{W}_\mathrm{ame}$ and pejoration $\mathcal{W}_\mathrm{pej}$}
    \label{tab:known_words_amelioration_and_pejoration}
\end{table}

\begin{figure*}[t]
\centering
\begin{minipage}[t]{1.0\columnwidth}
    \centering
    \includegraphics[width=1.0\columnwidth]{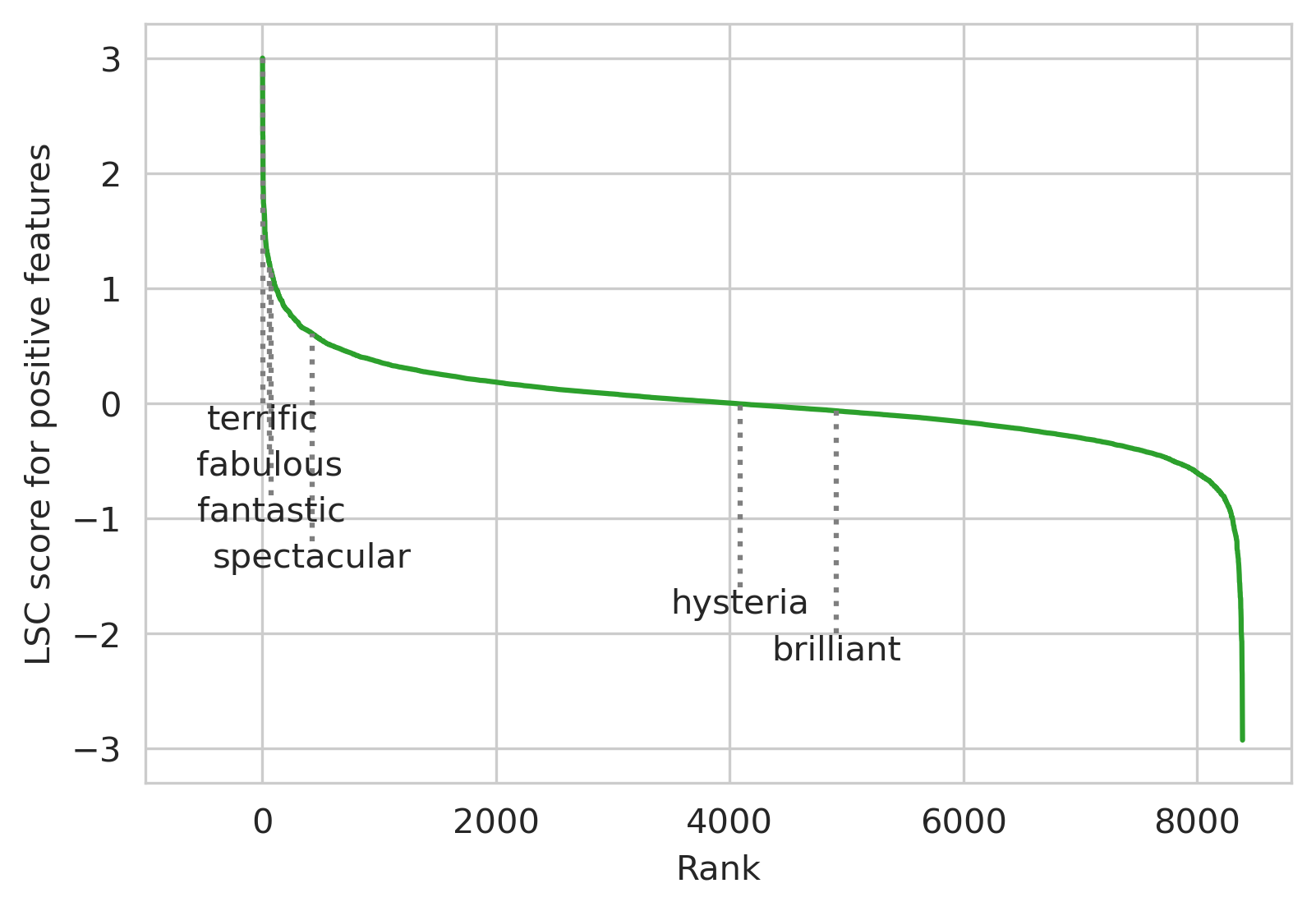}
    \caption{LSC scores for positive features}
    \label{fig:change_scores_pos}
\end{minipage}
\begin{minipage}[t]{1.0\columnwidth}
    \centering
    \includegraphics[width=1.0\columnwidth]{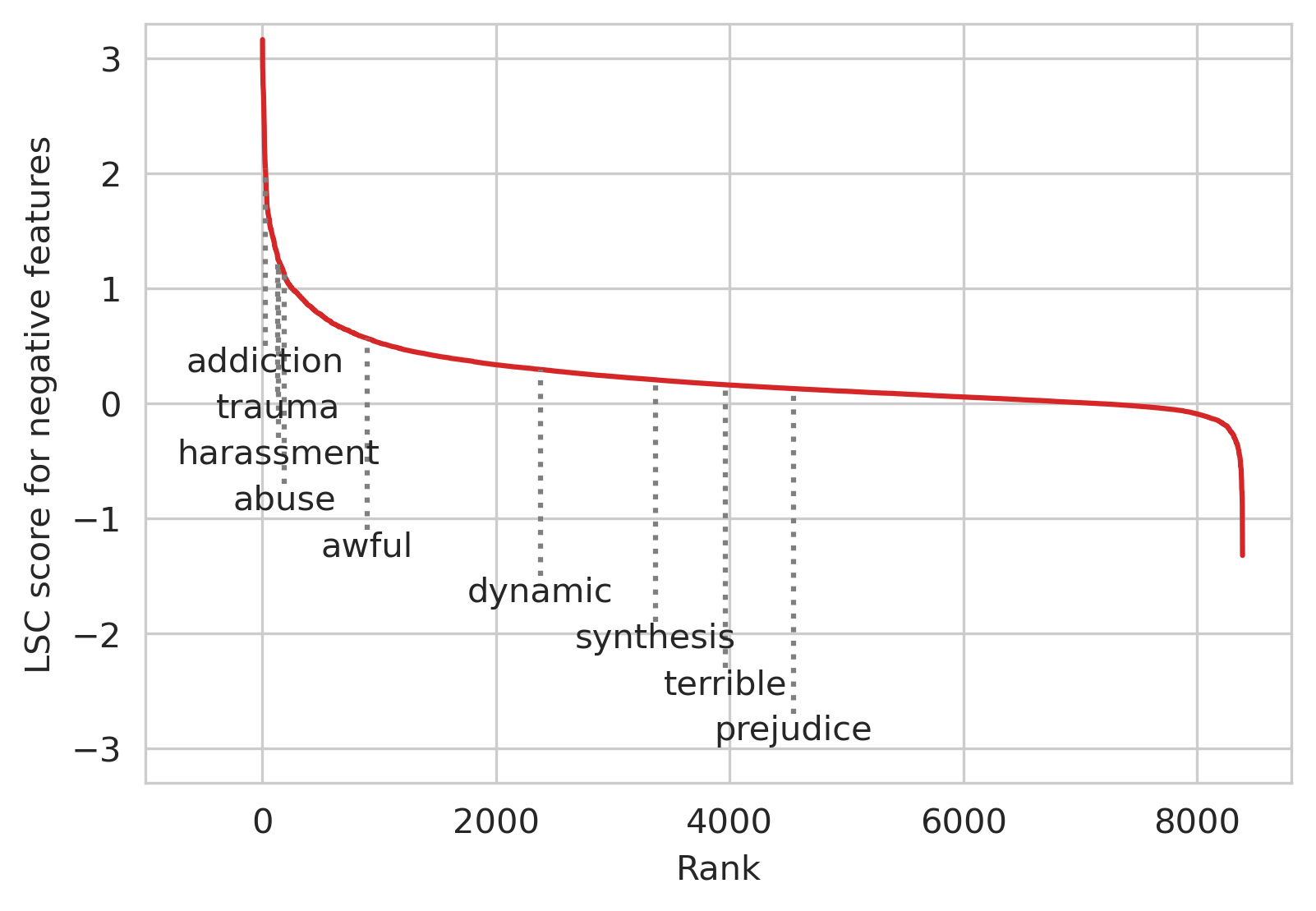}
    \caption{LSC scores for negative features}
    \label{fig:change_scores_neg}
\end{minipage}
\end{figure*}

\subsection{Methods}
First, we select the Binder features that are related to positive or negative meanings. 
Referring to \citet{Binder18052016}, the features related to positive meanings $\mathcal{I}_\mathrm{pos}$ are defined as \textit{Pleasant} and \textit{Happy}, while the features related to negative meanings $\mathcal{I}_\mathrm{neg}$ are defined as \textit{Pain}, \textit{Harm}, \textit{Unpleasant}, \textit{Sad}, \textit{Angry}, \textit{Disgusted}, and \textit{Fearful}. 
% These features are chosen because (i) they are highly correlated with the principal component PC9 (amelioration) or PC3 (pejoration), and (ii) their values are low within the eigenvectors of other PCs, which prevents other types of LSC from being considered. 
Indeed, some of these features indicate that the LSC type of PC9 and PC3 in Table \ref{tab:lsc_types} are amelioration and pejoration, respectively. 
Furthermore, \textit{Happy}, \textit{Sad}, \textit{Angry}, \textit{Disgusted}, and \textit{Fearful} are derived from the basic emotions proposed by \citet{Ekman01051992}, which are closely related to emotion analysis \citep{plaza-del-arco-etal-2024-emotion} in the field of NLP. 

Next, for each target word collected in Section \ref{subsec:analyzing_tyes_of_lsc-target_words}, a score indicating the degree of positive or negative lexical semantic change (called LSC score in this paper) is calculated as follows: 
\begin{equation}
    \mathrm{LSCS}(w, x) = \max_{i \in \mathcal{I}_x}\mathbf{v}_\mathrm{lsc}(w)[i], 
\end{equation}
where $\mathcal{I}_x$ is either $\mathcal{I}_\mathrm{pos}$ or $\mathcal{I}_\mathrm{neg}$. 
That is, the maximum value of the positive (or negative) features in the LSC vector is employed as the LSC score. 
Our motivation behind this definition is that a word should be recognized as amelioration or pejoration if one of the features in $\mathcal{I}_\mathrm{pos}$ or $\mathcal{I}_\mathrm{neg}$ increases significantly. 

Finally, we sort all the target words in order of their LSC scores and verify whether the words in $\mathcal{W}_\mathrm{ame}$ or $\mathcal{W}_\mathrm{pej}$ are highly ranked. 

While previous methods \citep{cook-stevenson-2010-automatically, goworek-dubossarsky-2024-toward} are specialized for detecting amelioration and pejoration, our approach can extend to identify words of other LSC types discovered in Section \ref{sec:analyzing_tyes_of_lsc}. 

\subsection{Results}

Figures \ref{fig:change_scores_pos} and \ref{fig:change_scores_neg} show the LSC scores for positive features $\mathrm{LSCS}(w, \mathrm{pos})$ and negative features $\mathrm{LSCS}(w, \mathrm{neg})$, respectively. 
Words changing in a positive direction (i.e., the LSC score is greater than zero) account for about half of the total, while words changing in a negative direction account for about 75\%. 
This indicates that words tend to change in a negative direction more than in a positive direction. 

Figure \ref{fig:change_scores_pos} shows that the rank of most known words of amelioration in Table \ref{tab:known_words_amelioration_and_pejoration} are relatively high. 
In particular, \textit{terrific} is ranked first. 
The OED and \citet{deWit_2021} denote that \textit{terrific} began to be used with a positive meaning in addition to a negative one in the late 19th century, and today it is mainly used with a positive meaning. 
On the other hand, the LSC scores for \textit{hysteria} and \textit{brilliant} are nearly zero. 
For \textit{hysteria}, no positive meaning similar to those shown by \citet{cook-stevenson-2010-automatically} are found in the OED and examples in the CCOHA. 
For \textit{brilliant}, according to the OED, this word originally meant \textit{shining} and acquired the metaphorical meaning of \textit{splendid} around 1739. 
This semantic shift was not captured because the LSC score is measured between periods of the 1910s and 2000s. 

Figure \ref{fig:change_scores_neg} indicates that the meaning of all the known words of pejoration in Table \ref{tab:known_words_amelioration_and_pejoration} are shifted in a negative direction. 
The words \textit{abuse}, \textit{addiction}, \textit{harassment}, and \textit{trauma}, which are suggested by \citet{Haslam02012016}, are ranked relatively high. 
According to \citet{Haslam02012016}, as the meanings of these words expand, people become more sensitive to their negative connotations. 
On the other hand, the ranks of some words in $\mathcal{W}_\mathrm{pej}$ are low. 
For \textit{prejudice}, the results are similar to those of \citet{vylomova-etal-2019-evaluation}, and unlike other words in \citet{Haslam02012016}, its meaning has not drastically shifted in a negative direction. 
For \textit{dynamic} and \textit{synthesis}, no negative meaning similar to those shown by \citet{cook-stevenson-2010-automatically} is found in the OED and examples in the CCOHA. 
For \textit{terrible}, since this word has only negative meaning, it is unlikely that its meaning will change in a more negative direction. 

% \begin{figure}[t]
% \centering
% \begin{minipage}[t]{0.35\columnwidth}
%     \centering
%     \includegraphics[width=1.0\columnwidth]{figures/embs_terrific.png}
%     \caption{T-SNE visualization for contextualized embeddings of \textit{terrific}}
%     \label{fig:embs_terrific}
% \end{minipage}
% \hfill
% \begin{minipage}[t]{0.6\columnwidth}
%     \centering
%     \includegraphics[width=1.0\columnwidth]{figures/sense_dist_terrific.png}
%     \caption{The distribution of usage types in the 1910s and 2000s of \textit{terrific}}
%     \label{fig:sense_dist_terrific}
% \end{minipage}
% \end{figure}

% An additional analysis is carried out for the LSC of \textit{terrific}. 
% Figure \ref{fig:embs_terrific} shows a visualization of the contextualized embeddings, while Figure \ref{fig:sense_dist_terrific} shows the distribution of usage types for each period. 
% In these figures, the colors blue and orange indicate the meanings ``amazing'' and ``causing terror'', respectively. 
% These results indicate that the different meanings of terrific are appropriately distinguished in a vector space, and that its main meaning has changed from negative to positive. 

To sum up, these results demonstrate the effectiveness of our method in the detection of amelioration and pejoration. 

\section{Conclusion}

This study proposed a novel method to improve the interpretability of LSC by mapping the semantic space of the pre-trained language model to the neurobiological space. 
In the experiments designed to estimate the degree of LSC, our method demonstrated better performance than the baseline methods that did not map the semantic spaces. 
By leveraging the high interpretability of our method, we discovered interesting types of LSC that had not been identified previously. 
Additionally, in the detection of amelioration and pejoration, our method assigned appropriate LSC scores for words, which evaluated how their meanings changed positively or negatively. 
In the future, we plan to apply our method to detect words of other types of LSC. 

\clearpage

\section*{Limitations}

In this study, we analyzed several LSC types from the perspective of the Binder features. 
On the other hand, according to \citet{1360013171975918848}, there are different types of LSC, such as metaphorization, metonymization, narrowing, and generalization. 
The method proposed in this paper might struggle to capture these LSC types because there is no clear correlation between the Binder features and them. 
Therefore, it is necessary to extend the current method or adopt new methods of representation (e.g., representing the meaning of a word in a sentence with box embeddings \citep{oda-etal-2024-learning}). 

In addition, it is necessary to increase the number of target words. 
In our method, words that are not included in the vocabulary of the tokenizer of pre-trained language models are outside the scope of the analysis, resulting in failure to capture the LSC of those words. 
Even when a word is split into multiple subwords, contextualized embeddings should be obtained, for example, by taking an average vector of the contextualized embeddings of these subwords \citep{montariol-etal-2021-scalable}. 

\bibliography{anthology1,anthology2,custom}

\appendix

\section{Distributions of the usage types}
\label{sec:distributions_usage_types}

\input{figure_usage_type_dist.tex}

The distributions of the usage types for several representative words in Table \ref{tab:lsc_types} are shown in Figures \ref{fig:sense_dist_appendix1} and \ref{fig:sense_dist_appendix2}. 

\end{document}

%% file: figure_usage_type_dist.tex
\begin{figure*}[t]
    \centering
    \begin{subfigure}[t]{0.45\linewidth}
        \includegraphics[width=\linewidth]{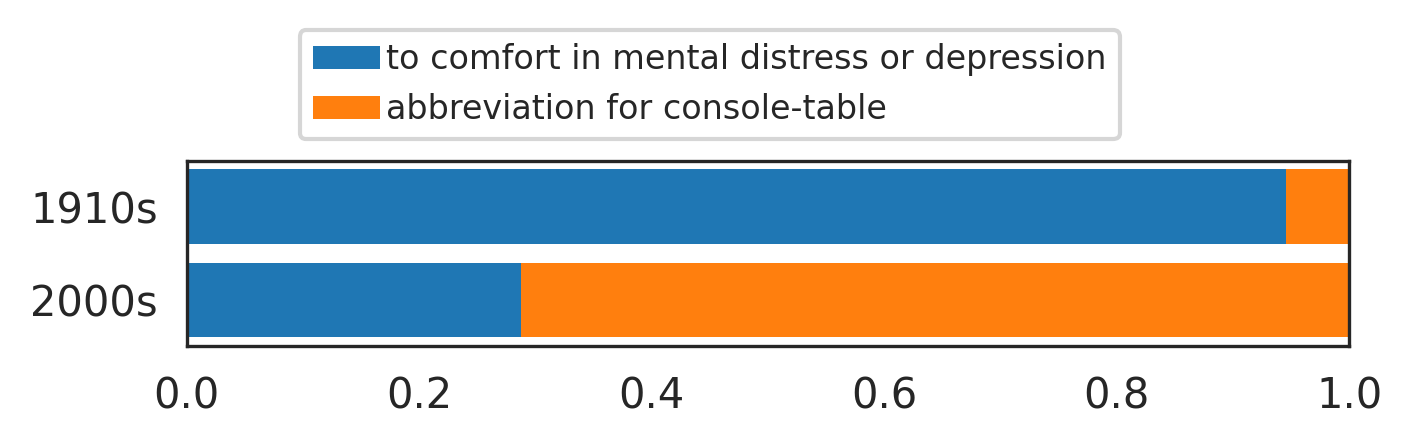}
        \caption{console [PC1, Artifact, $\uparrow$]}
        \label{fig:sense_dist_appendix_console}
    \end{subfigure}
    \begin{subfigure}[t]{0.45\linewidth}
        \includegraphics[width=\linewidth]{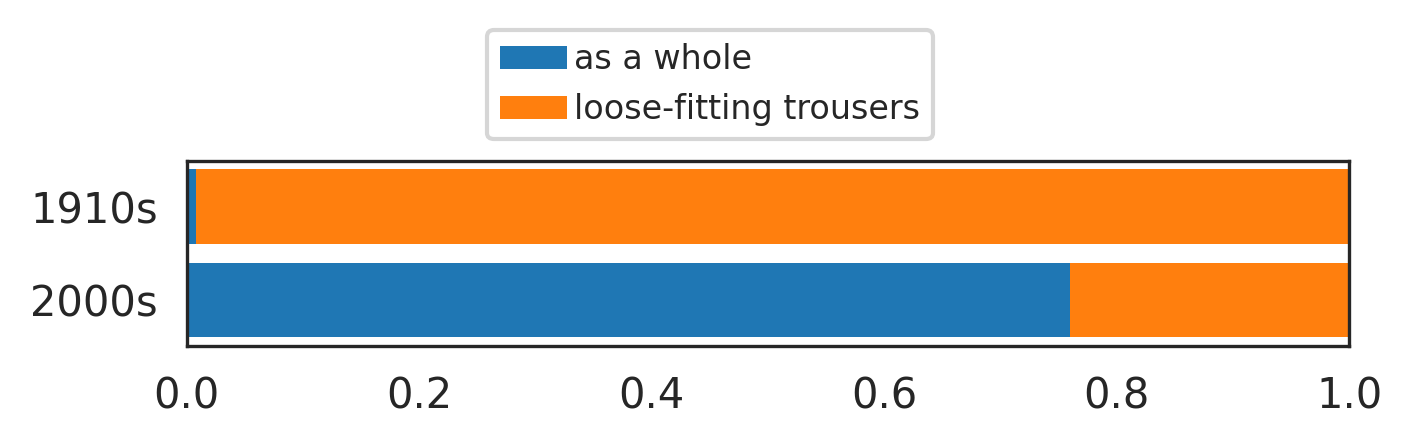}
        \caption{overall [PC1, Artifact, $\downarrow$]}
        \label{fig:sense_dist_appendix_overall}
    \end{subfigure}

    \bigskip
    
    \begin{subfigure}[t]{0.45\linewidth}
        \includegraphics[width=\linewidth]{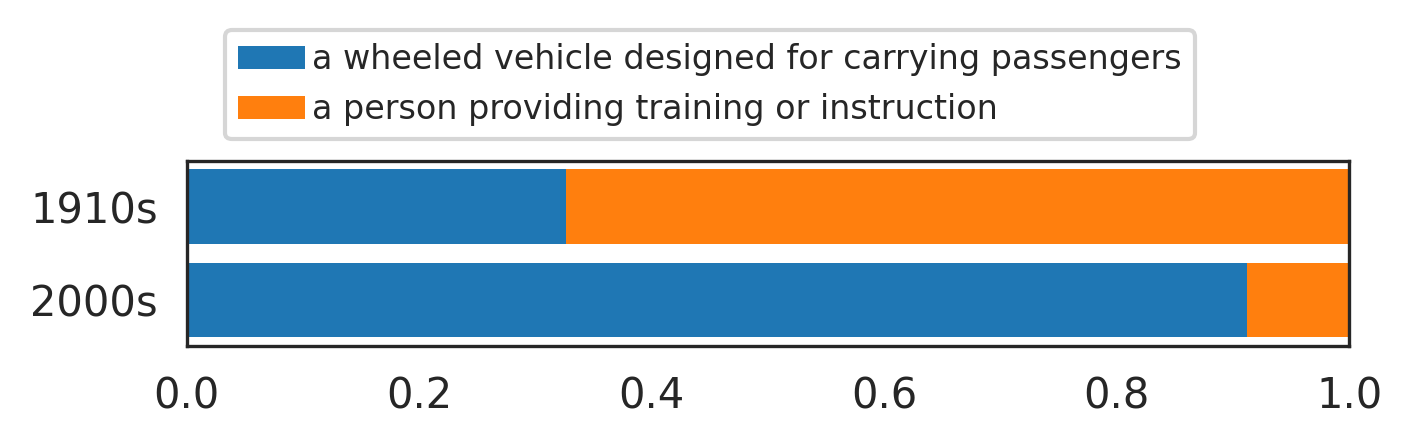}
        \caption{coach [PC2, Human, $\uparrow$]}
        \label{fig:sense_dist_appendix_coach}
    \end{subfigure}
    \begin{subfigure}[t]{0.45\linewidth}
        \includegraphics[width=\linewidth]{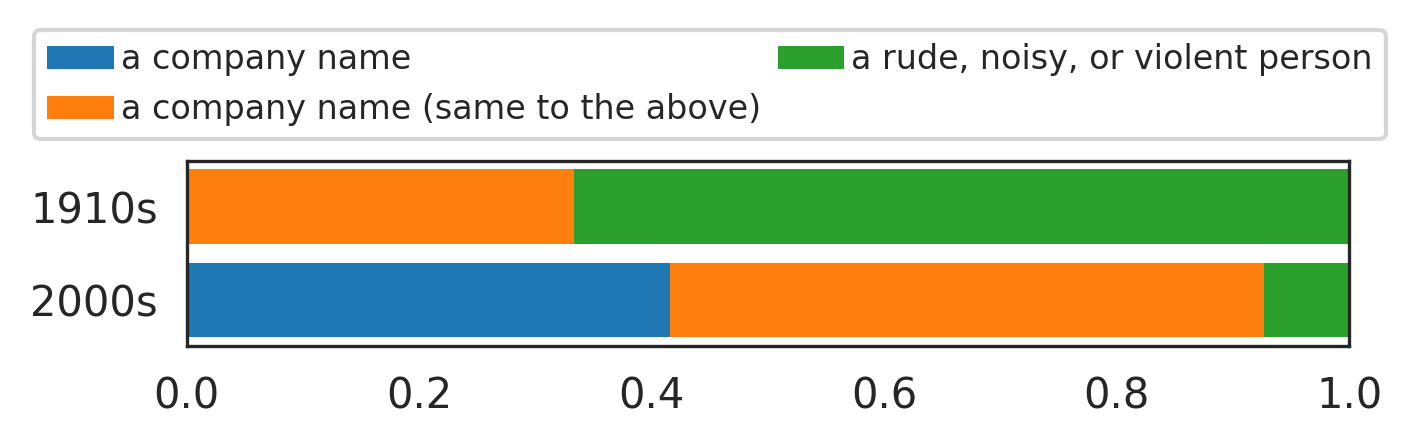}
        \caption{yahoo [PC2, Human, $\downarrow$]}
        \label{fig:sense_dist_appendix_yahoo}
    \end{subfigure}

    \bigskip
    
    \begin{subfigure}[t]{0.45\linewidth}
        \includegraphics[width=\linewidth]{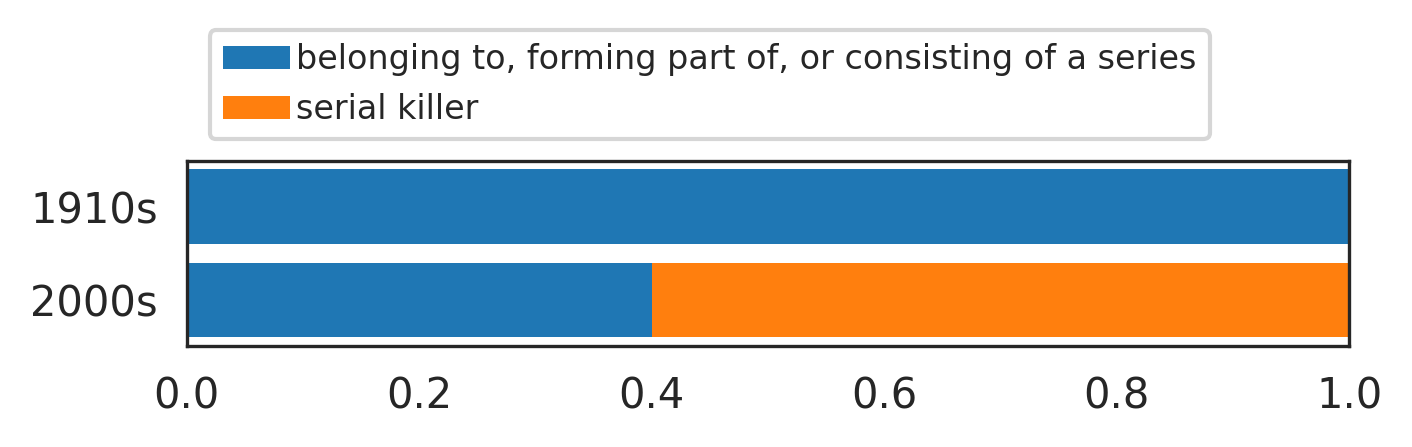}
        \caption{serial [PC3, Negative Meaning, $\uparrow$]}
        \label{fig:sense_dist_appendix_serial}
    \end{subfigure}
    \begin{subfigure}[t]{0.45\linewidth}
        \includegraphics[width=\linewidth]{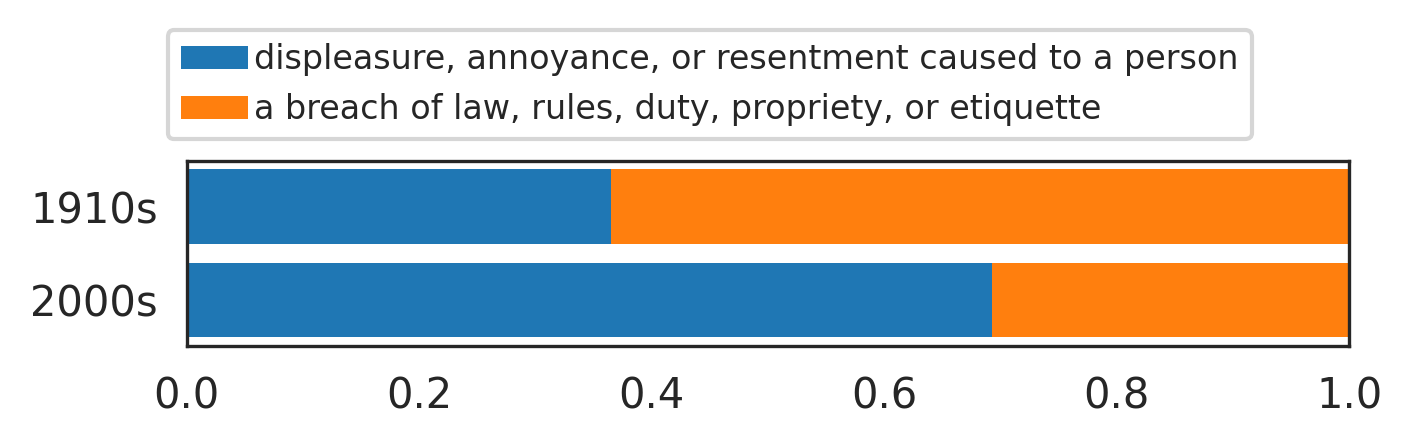}
        \caption{offence [PC3, Negative Meaning, $\downarrow$]}
        \label{fig:sense_dist_appendix_offence}
    \end{subfigure}

    \bigskip
    
    \begin{subfigure}[t]{0.45\linewidth}
        \includegraphics[width=\linewidth]{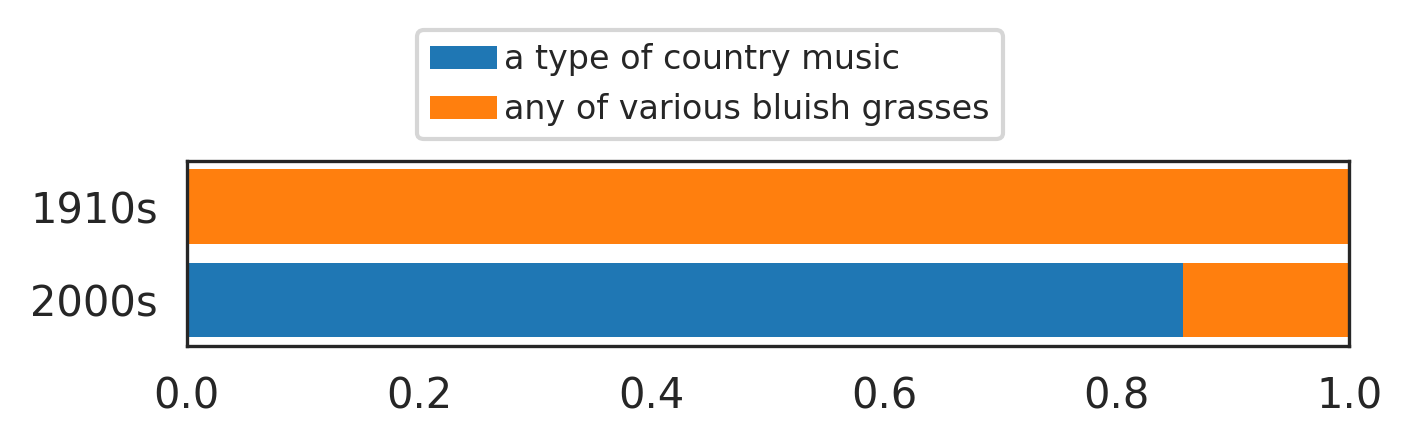}
        \caption{bluegrass [PC4, Sound, $\uparrow$]}
        \label{fig:sense_dist_appendix_bluegrass}
    \end{subfigure}
    \begin{subfigure}[t]{0.45\linewidth}
        \includegraphics[width=\linewidth]{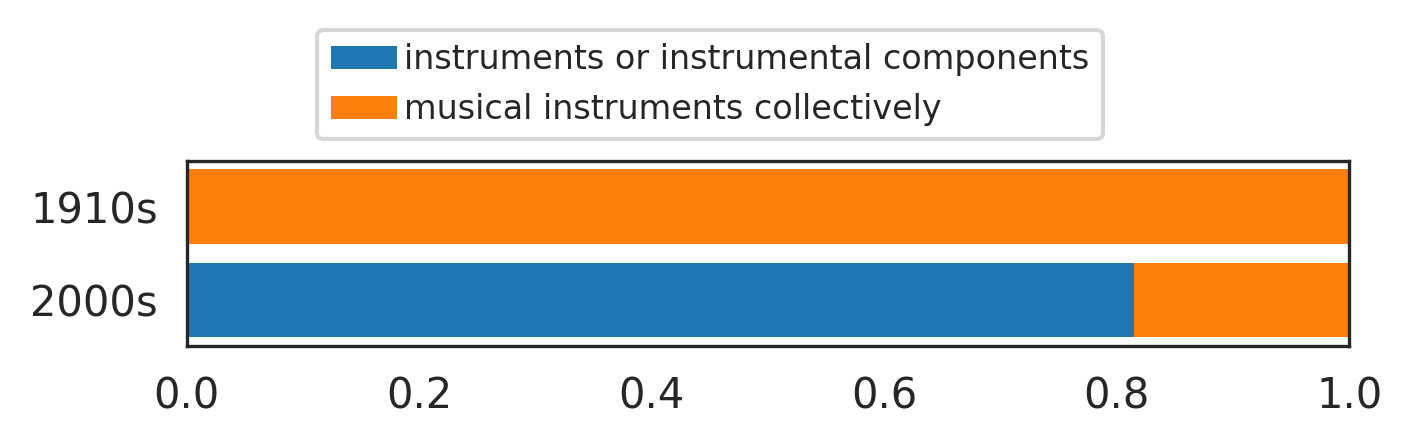}
        \caption{instrumentation [PC4, Sound, $\downarrow$]}
        \label{fig:sense_dist_appendix_instrumentation}
    \end{subfigure}

    \bigskip
    
    \begin{subfigure}[t]{0.45\linewidth}
        \includegraphics[width=\linewidth]{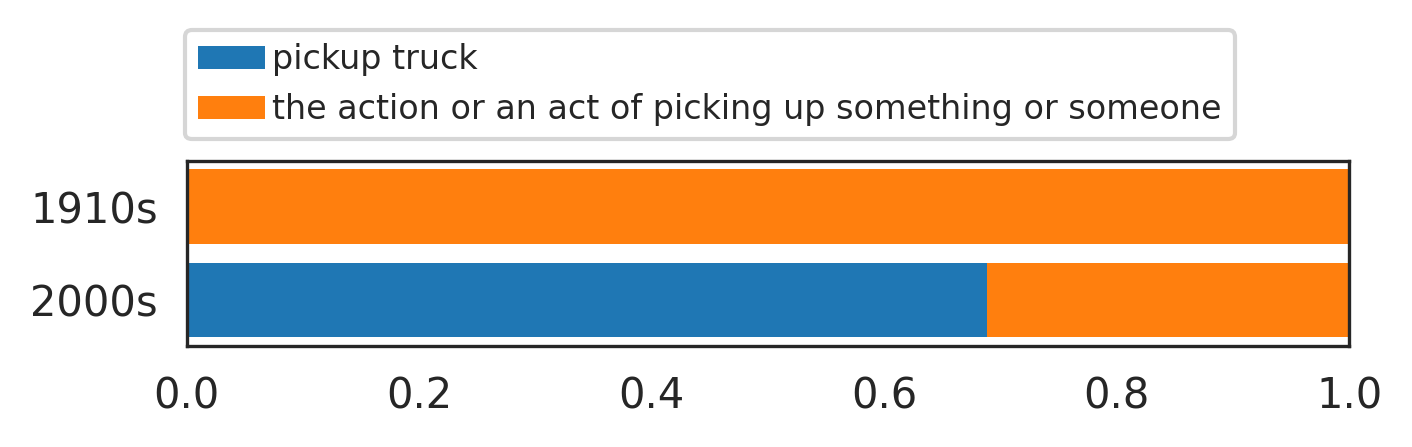}
        \caption{pickup [PC5, Transportation, $\uparrow$]}
        \label{fig:sense_dist_appendix_pickup}
    \end{subfigure}
    \begin{subfigure}[t]{0.45\linewidth}
        \includegraphics[width=\linewidth]{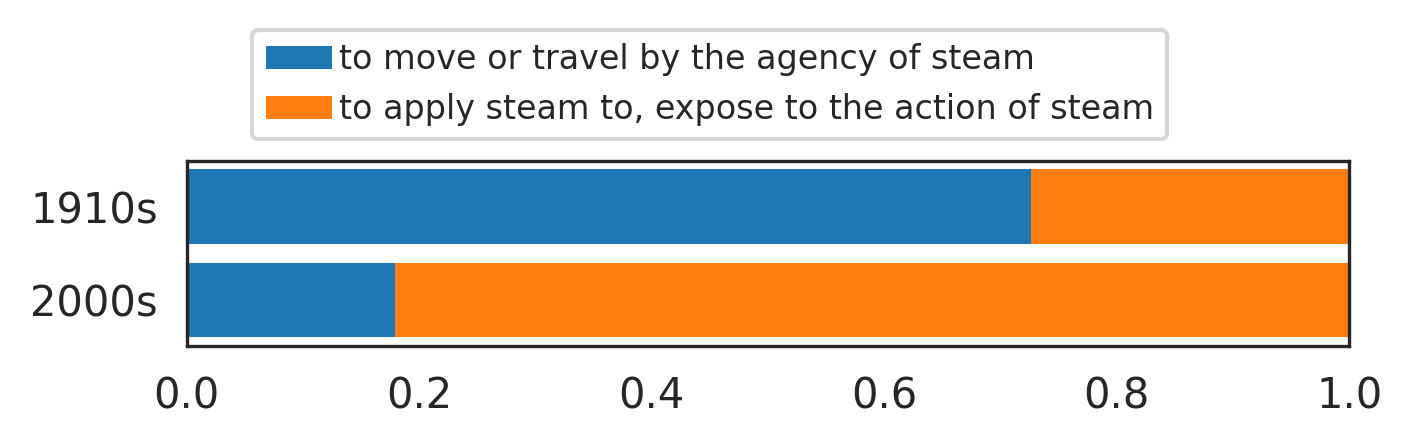}
        \caption{steamed [PC5, Transportation, $\downarrow$]}
        \label{fig:sense_dist_appendix_steamed}
    \end{subfigure}

    \bigskip
    
    \begin{subfigure}[t]{0.45\linewidth}
        \includegraphics[width=\linewidth]{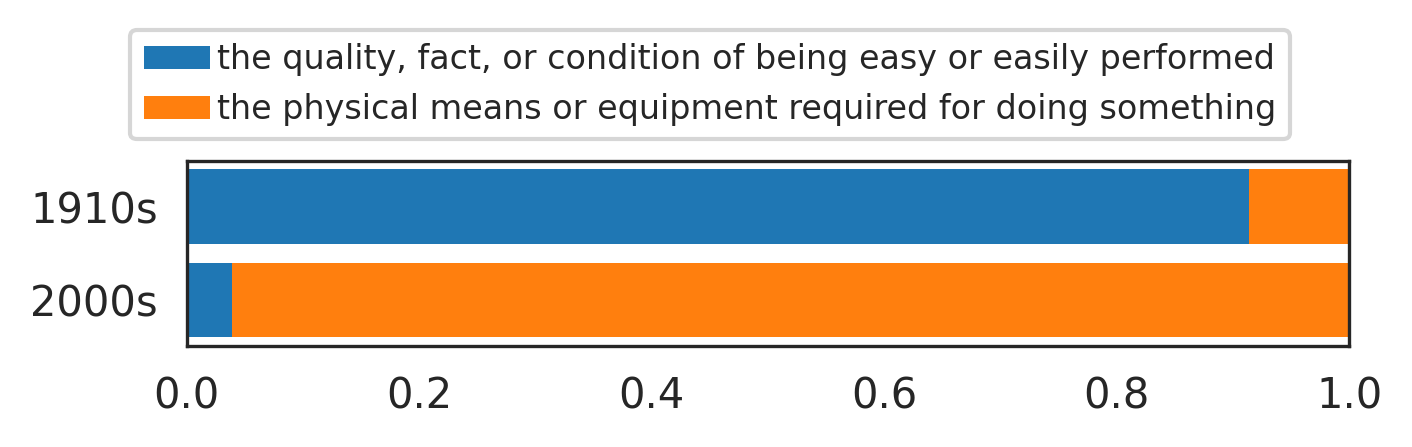}
        \caption{facility [PC6, Place, $\uparrow$]}
        \label{fig:sense_dist_appendix_facility}
    \end{subfigure}
    \begin{subfigure}[t]{0.45\linewidth}
        \includegraphics[width=\linewidth]{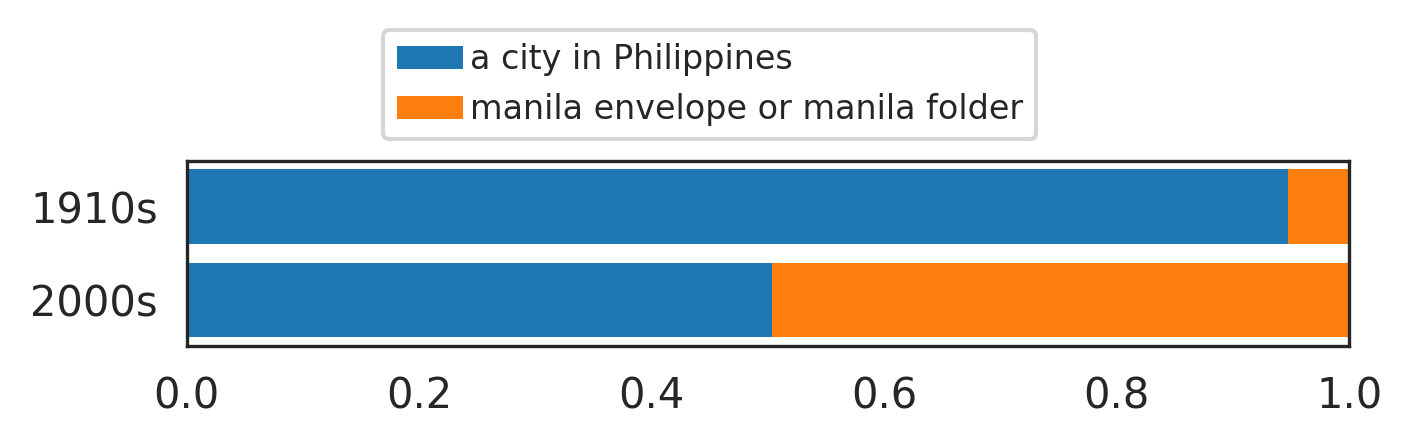}
        \caption{manila [PC6, Place, $\downarrow$]}
        \label{fig:sense_dist_appendix_manila}
    \end{subfigure}

    \bigskip
    
    \caption{Distributions of usage types of representative words. These words are excerpted from Table~\ref{tab:lsc_types}. An ID of a principal component, an LSC type label, and an arrow indicating the direction of semantic change of each word are in parentheses.}
    \label{fig:sense_dist_appendix1}
\end{figure*}

\begin{figure*}[t]
    \centering
    \begin{subfigure}[t]{0.45\linewidth}
        \setcounter{subfigure}{12}
        \includegraphics[width=\linewidth]{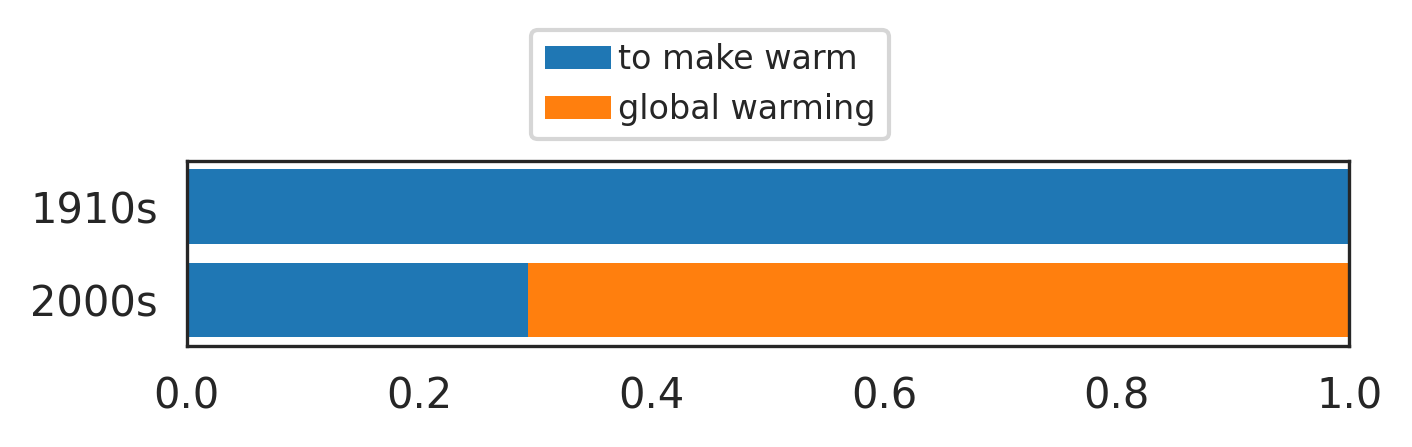}
        \caption{warming [PC7, Social, $\uparrow$]}
        \label{fig:sense_dist_appendix_warming}
    \end{subfigure}
    \begin{subfigure}[t]{0.45\linewidth}
        \includegraphics[width=\linewidth]{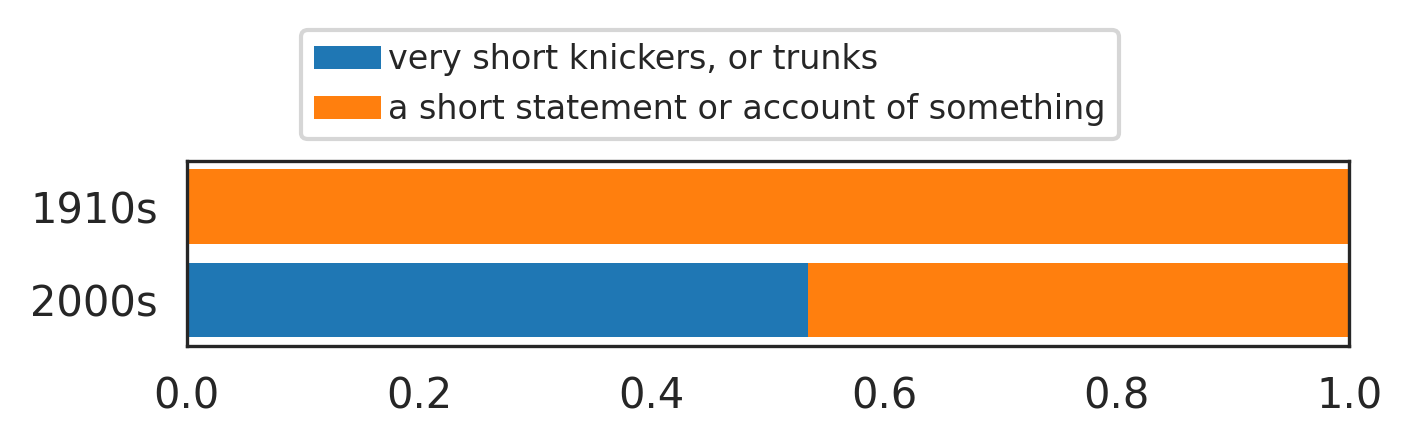}
        \caption{briefs [PC7, Social, $\downarrow$]}
        \label{fig:sense_dist_appendix_briefs}
    \end{subfigure}

    \bigskip

    \begin{subfigure}[t]{0.45\linewidth}
        \includegraphics[width=\linewidth]{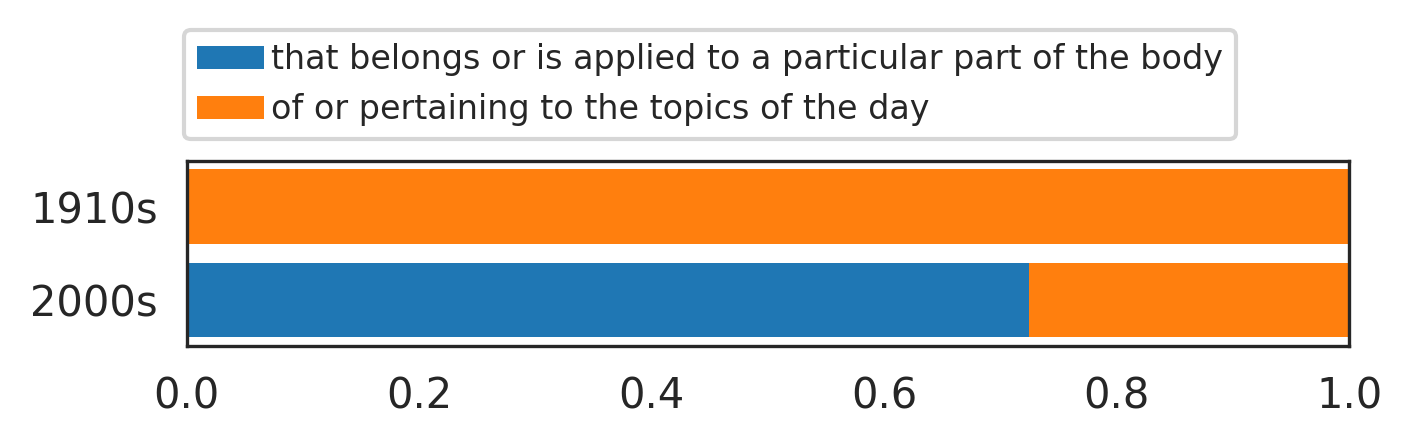}
        \caption{topical [PC8, Familiar Thing, $\uparrow$]}
        \label{fig:sense_dist_appendix_topical}
    \end{subfigure}
    \begin{subfigure}[t]{0.45\linewidth}
        \includegraphics[width=\linewidth]{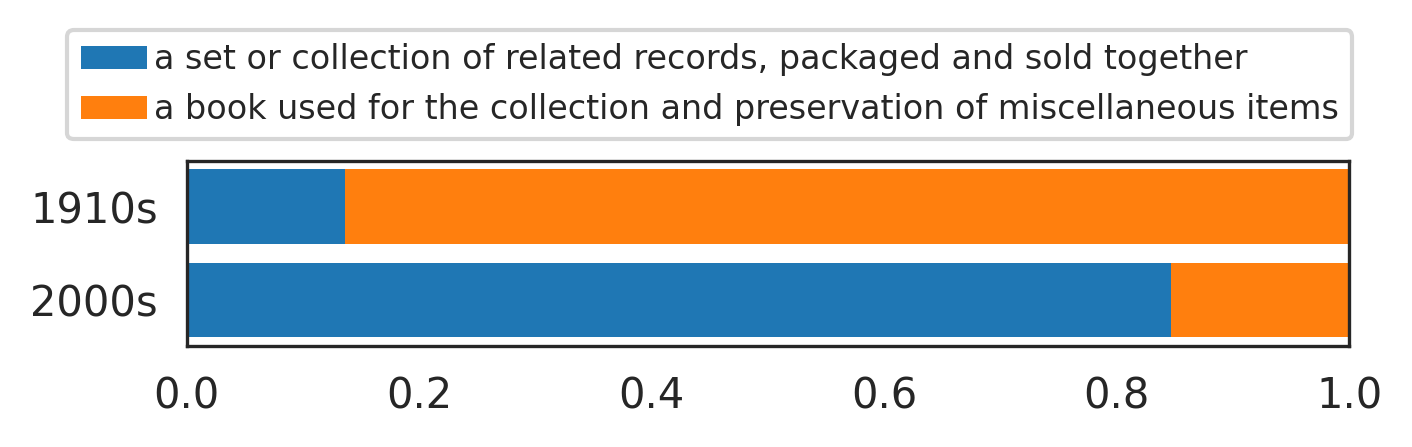}
        \caption{album [PC8, Familiar Thing, $\downarrow$]}
        \label{fig:sense_dist_appendix_album}
    \end{subfigure}

    \bigskip
    
    \begin{subfigure}[t]{0.45\linewidth}
        \includegraphics[width=\linewidth]{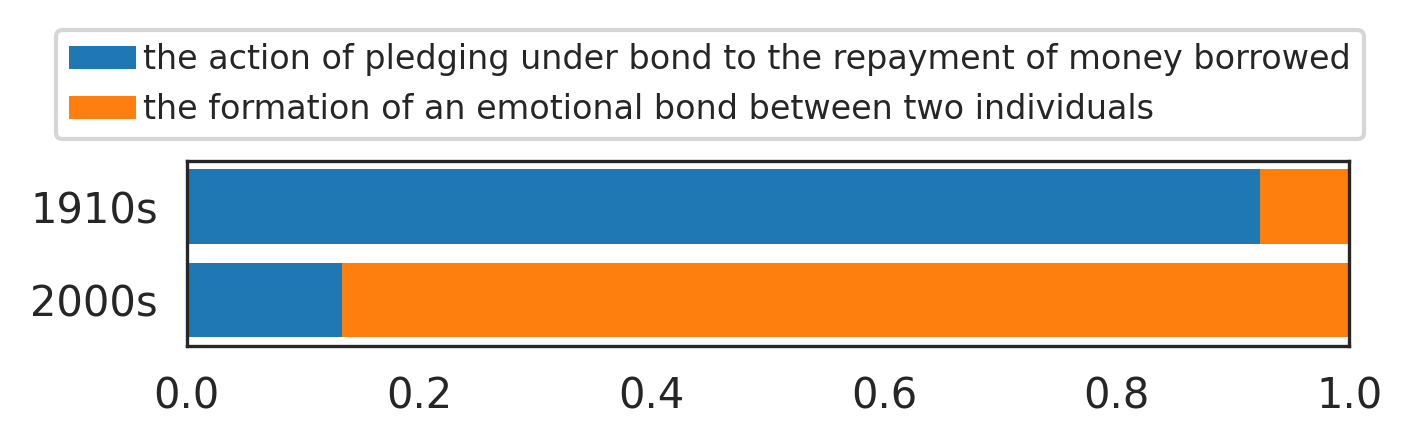}
        \caption{bonding [PC9, Positive Meaning, $\uparrow$]}
        \label{fig:sense_dist_appendix_bonding}
    \end{subfigure}
        \begin{subfigure}[t]{0.45\linewidth}
        \includegraphics[width=\linewidth]{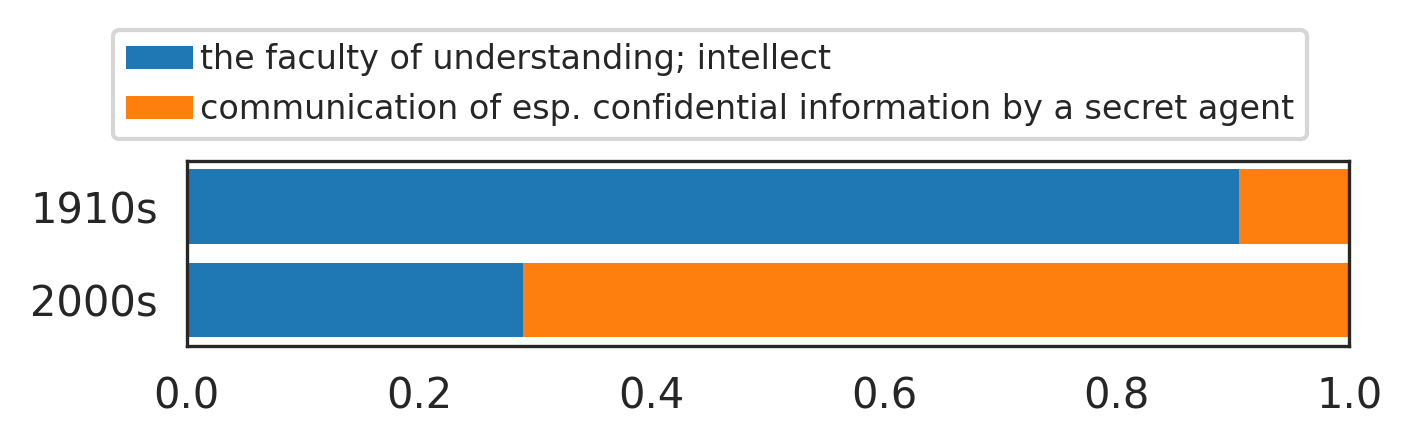}
        \caption{intelligence [PC9, Positive Meaning, $\downarrow$]}
        \label{fig:sense_dist_appendix_intelligence}
    \end{subfigure}

    \bigskip
    
    \begin{subfigure}[t]{0.45\linewidth}
        \includegraphics[width=\linewidth]{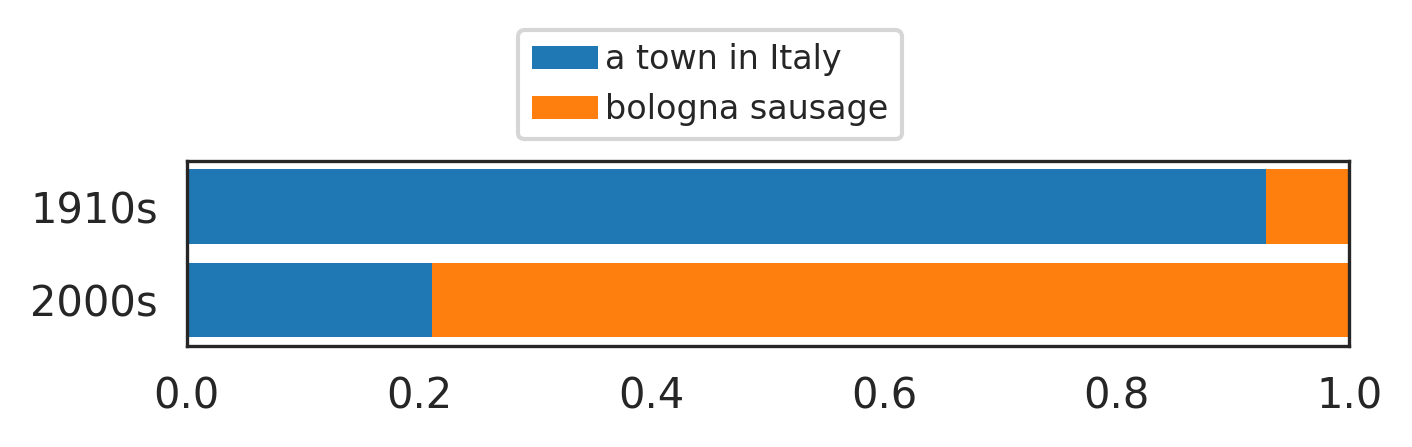}
        \caption{bologna [PC10, Food, $\uparrow$]}
        \label{fig:sense_dist_appendix_bologna}
    \end{subfigure}
    \begin{subfigure}[t]{0.45\linewidth}
        \includegraphics[width=\linewidth]{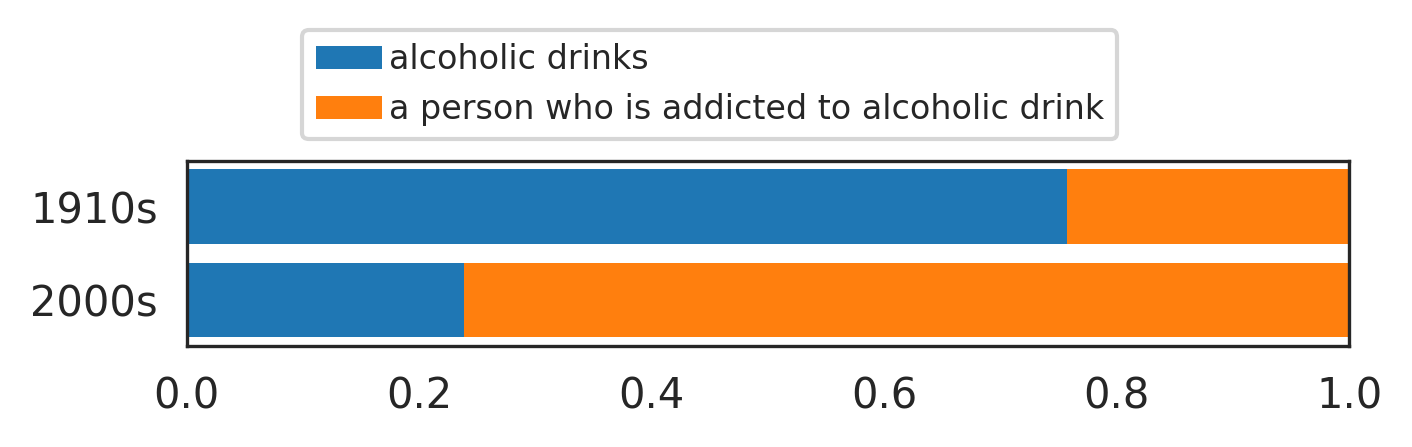}
        \caption{alcoholic [PC10, Food, $\downarrow$]}
        \label{fig:sense_dist_appendix_alcoholic}
    \end{subfigure}

    \bigskip
    
    \caption{Distributions of usage types of representative words (cont.)}
    \label{fig:sense_dist_appendix2}
\end{figure*}